\documentclass{article}

\usepackage{microtype}
\usepackage{graphicx}
\usepackage{subcaption}
\usepackage{booktabs} 

\usepackage{hyperref}

\usepackage[preprint]{icml2026}

\usepackage{amsmath}
\usepackage{amssymb}
\usepackage{mathtools}
\usepackage{amsthm}

\usepackage[capitalize,noabbrev]{cleveref}

\newtheorem{assumption}{Assumption}

\newtheorem{definition}{Definition}

\usepackage[textsize=tiny]{todonotes}

\usepackage{amsmath}
\usepackage{amssymb}
\usepackage{mathtools}
\usepackage{amsthm}
\usepackage{bbm}
\usepackage{multirow}
\usepackage{mathrsfs}
\usepackage{graphics}
\usepackage{wrapfig}
\usepackage{epstopdf}
\usepackage{pdfpages}
\usepackage{diagbox}
\usepackage{ulem}
\usepackage{MnSymbol}
\usepackage{pifont} 

\usepackage[table]{xcolor}
\usepackage{tabularx}
\usepackage{diagbox}
\usepackage{makecell}
\usepackage{diagbox}

\icmltitlerunning{Are We Evaluating the Edit Locality of LLM Model Editing Properly?}

\begin{document}

\twocolumn[
  \icmltitle{Are We Evaluating the Edit Locality of LLM Model Editing Properly?}



  \icmlsetsymbol{equal}{*}

  \begin{icmlauthorlist}
    \icmlauthor{Wei Liu}{aaa}
    \icmlauthor{Haomei Xu}{bbb}
    \icmlauthor{Hongkai Liu}{aaa}
    \icmlauthor{Zhiying Deng}{ddd}
    \icmlauthor{Ruixuan Li}{bbb}
    \icmlauthor{Heng Huang}{eee}
    \icmlauthor{Yee Whye Teh}{fff}
    \icmlauthor{Wee Sun Lee}{aaa}
  \end{icmlauthorlist}

  \icmlaffiliation{aaa}{National University of Singapore}
  \icmlaffiliation{bbb}{Huazhong University of Science and Technology}
  \icmlaffiliation{ddd}{Central China Normal University}
  \icmlaffiliation{eee}{University of Maryland, College Park}
  \icmlaffiliation{fff}{Oxford}
  \icmlcorrespondingauthor{Zhiying Deng}{zhiyingdzy@gmail.com}
  \icmlcorrespondingauthor{Ruixuan Li}{rxli@hust.edu.cn}
  \icmlcorrespondingauthor{Wee Sun Lee}{dcsleews@nus.edu.sg}
  \icmlkeywords{Machine Learning, ICML}

  \vskip 0.3in
]



\printAffiliationsAndNotice{}  

\normalem

\begin{abstract}
Model editing has recently emerged as a popular paradigm for efficiently updating knowledge in LLMs. 
A central desideratum of updating knowledge is to balance editing efficacy, i.e., the successful injection of target knowledge, and specificity (also known as edit locality), i.e., the preservation of existing non-target knowledge. 
However, we find that existing specificity evaluation protocols are inadequate for this purpose. 
We systematically elaborated on the three fundamental issues it faces.
Beyond the conceptual issues, we further empirically demonstrate that existing specificity metrics are weakly correlated with the strength of specificity regularizers. We also find that current metrics lack sufficient sensitivity, rendering them ineffective at distinguishing the specificity performance of different methods. Finally, we propose a constructive evaluation protocol. Under this protocol, the conflict between open-ended LLMs and the assumption of determined answers is eliminated, query-independent fluency biases are avoided, and the evaluation strictness can be smoothly adjusted within a near-continuous space. 
Experiments across various LLMs, datasets, and editing methods show that metrics derived from the proposed protocol are more sensitive to changes in the strength of specificity regularizers and exhibit strong correlation with them, enabling more fine-grained discrimination of different methods’ knowledge preservation capabilities.
\end{abstract}

\section{Introduction}
Large language models (LLMs) have demonstrated remarkable capabilities across a wide range of natural language processing tasks. However, their knowledge is constrained by static training data, causing them to encode outdated, incomplete, or even incorrect facts. Updating such knowledge can be computationally expensive and often impractical given the scale of modern LLMs.

\begin{figure}[t]
\centering
    \includegraphics[width=0.99\columnwidth]{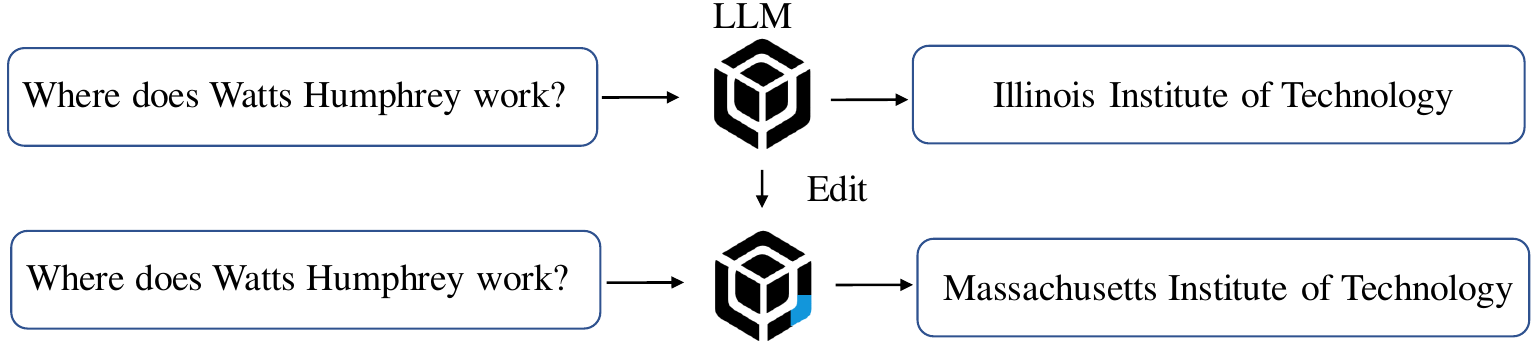}
    \caption{An example of LLM editing: updating the incorrect knowledge with modifying only a small set of parameters.}
    \label{fig: edit_example}
    \vspace{-15pt}
\end{figure}

To address this limitation, model parameter editing has been proposed as an alternative for updating specific pieces of knowledge in pretrained models. By precisely locating and editing a small subset of parameters (see Figure \ref{fig: edit_example}), parameter editing \cite{rome} offers a promising balance between efficiency and effectiveness, and has recently emerged as one of the mainstream popular paradigms.

A core desideratum in model editing literature lies in balancing two competing goals. On the one hand, an editing should achieve high efficacy, meaning that the target knowledge is effectively updated. On the other hand, it should maintain strong specificity, ensuring that other model behavior remains unaffected. To get a good trade-off between efficacy and specificity, we need to first be able to properly evaluate them.   However, we find that the evaluation about the specificity in the current literature can be problematic. 

Existing specificity evaluation protocols \citep{rome,memit} typically rely on a set of queries paired with pre-defined determined ground-truth answer. Specificity is then measured by the post-edit model’s probability on generating this definite ground-truth answer (e.g., teacher-forcing token match, comparison with contrastive candidates, etc). However, we argue that commonly used specificity evaluation protocols are ill-suited to capture the true extent of behavioral change induced by model editing. 

Through systematic analysis, we show that these evaluations may overlook several fundamental issues. First, they rely on pre-defined definite ground-truth answers as proxies for the model’s pre-edit behavior. Pre-defined answers may not necessarily agree with the model's internal knowledge, and ``definite'' contrasts with the open-ended property of modern LLMs as an LLM's  response to a query is not restricted to a definite token string (e.g., ``Massachusetts Institute of Technology'' \emph{vs} ``MIT'').
Thus, the model’s original responses often do not align with such annotations, leading to substantial misalignment that is unrelated to the edit itself. 
Second, many specificity metrics (e.g., contrastive candidate comparison and teacher-forcing token match ratio) are susceptible to query-independent language priors: scores may be dominated by answer fluency or token-level co-occurrence patterns rather than by whether the underlying knowledge associated with the query has been preserved (e.g., after seeing the first answer token ``Massachusetts'', the model may have high probability to generate ``Institute of Technology'', regardless of the query). Third, specificity is inherently a continuous notion, yet ground-truth-based evaluation can only provide highly discrete metrics. This discretization renders metrics either overly strict, penalizing benign surface-form variations, or overly permissive, failing to detect moderate but meaningful shifts.

We then use real-world experiments to show how often these conceptual issues can occur in practice ($\S$\ref{sec: verification of multi issues}). Moreover, we argue that these issues are not merely conceptual; they can severely confound our assessment of the relative capabilities of different methods. We further empirically demonstrate that widely used specificity metrics exhibit weak correlation with the strength of specificity-preserving regularizers widely employed in current locate-then-edit methods.\footnote{Since locate-then-edit methods share a uniform design of explicit specificity regularizers, they serve as a natural choice for studying specificity evaluation.} 
Moreover, they often lack sufficient sensitivity to changes in regularization strength, making it difficult to distinguish editing methods with genuinely different impacts on unrelated knowledge. As a result, current specificity evaluations may produce misleading conclusions, such as suggesting that certain methods achieve large gains in efficacy with negligible specificity loss.

To address these limitations, we propose a constructive ground-truth-free evaluation protocol for specificity. Instead of comparing the edited model’s outputs to external reference answers, our approach directly measures the behavioral deviation between the model before and after editing. We instantiate this protocol using distributional and ranking-based metrics derived from model query logits. By operating at the level of output distributions, this protocol avoids reliance on determined answers, mitigates query-independent fluency biases, and represents specificity as a continuous quantity that can be measured at varying resolutions. 

Extensive experiments show that specificity metrics derived from our protocol are substantially more sensitive to changes in specificity regularization strength and exhibit significantly stronger alignment with specificity-oriented regularizers than existing metrics. Furthermore, they enable more fine-grained and stable comparison of different editing methods in terms of specificity preservation.

In summary, the contributions include:
\begin{itemize}
    \item We provide a systematic demonstration of limitations in existing specificity evaluation protocols for model editing, highlighting their contraction to open-ended LLM settings, with not only multi-level conceptual analysis but also fine-grained empirical verifications.
    \item We propose a human-interpretable, behavior-based evaluation protocol that measures specificity as continuous deviation between pre- and post-edit model behavior. Aside from being continuously adjustable, the proposed metric also preserves a percentage-based scale that eases the difficulty of human intuition.
    \item We empirically demonstrate that the proposed metrics are more sensitive and better aligned with the strength of specificity-preserving regularizers, enabling more reliable evaluation of model editing methods.
\end{itemize}
We note that this is an evaluation paper rather than a method paper, and the contribution type does not belong to proposing a novel method. Instead, the primary contribution is the systematic demonstration of the reasons why existing evaluation is problematic and why our proposed metrics work.

\section{Preliminaries of Model Editing}

\textbf{Notation}. 
Let $f_{\theta}$ denote a pre-trained LLM parameterized by $\theta$. 
After editing, the updated model is denoted as $f_{\theta^*}$. 
We define the target knowledge set as
\begin{equation}
   S^*=\{(q_i, y_i)\}_{i=1}^n,
\end{equation}
where $q_i$ is an input query that triggers a specific piece of factual knowledge (e.g., \textit{``Where does Watts Humphrey work''}), $y_i$ is the desired output after editing (e.g., \textit{``Massachusetts Institute of Technology''}), and $n$ denotes the number of knowledge items to be updated.

To encourage the preservation of other knowledge, a representative set
\begin{equation}
   S=\{(q_j, y_j)\}_{j=n+1}^{n+u}
\end{equation}
is typically sampled from a background corpus such as Wikipedia. 
Notably, $y_j$ are
obtained with $y_j=f_\theta(q_j)$, rather than really sampled from Wikipedia.

\textbf{General editing objective}. 
The overall goal of knowledge editing is to update specific factual associations while minimally affecting the model’s behavior on other inputs. 
This objective can be formulated as:
\begin{equation}\label{eqa: overall obj}
\begin{aligned}
    \theta^* = \mathop{\arg\min}_{\hat{\theta}}
    (
    & \sum_{i=1}^n \mathcal{L}_1(f_{\hat{\theta}}(q_i), y_i)
   \\& + \lambda \sum_{j=n+1}^{n+u} \mathcal{L}_2(f_{\hat{\theta}}(q_j), y_j)
    ),
\end{aligned}
\end{equation}
where $\mathcal{L}_1$ and $\mathcal{L}_2$ denote loss functions for enforcing the target edit and preserving other knowledge, respectively, and $\lambda$ controls the trade-off between the two objectives.

\textbf{Locate-then-edit paradigm}. 
Direct optimization of Eq. \ref{eqa: overall obj} using gradient descent is computationally expensive, because $u$ is usually very large to cover a wide range of non-target knowledge.
To address this challenge, prior work proposes the \emph{locate-then-edit} (LTE) paradigm \citep{rome,memit} that gives a succinct solution for it.
LTE first identifies where a piece of factual knowledge is stored in the model and then apply targeted parameter updates. A widely adopted approach is \emph{causal tracing} \citep{rome}, which identifies:
\begin{itemize}
    \item \textbf{Decisive token}: the token whose hidden representation most strongly determines the factual output, often corresponding to the final token of the subject entity in the input.
    \item \textbf{Decisive layer}: the model layer at which intervening on the hidden state of the decisive token most effectively steers the model toward the desired edit target, often corresponding to an MLP layer.
\end{itemize}
By intervening only on the hidden state of the decisive token at the decisive layer, these methods can selectively overwrite specific factual associations with high efficiency and precision. 
Due to its simplicity and strong empirical performance, this paradigm has become one of the mainstreams in the knowledge editing literature.

Despite differences in implementation, many locate-then-edit methods can be expressed as variants of the following optimization problem:
\begin{equation}\label{eq: memit}
\begin{aligned}
    \Delta^* = \mathop{\arg\min}_{\Delta}
    (&\sum_{i=1}^n \left\| (W + \Delta) k_i - m_i \right\|^2 \\
    &+ \lambda \sum_{j=n+1}^{n+u} \left\| \Delta k_j \right\|^2)
   , \\
    W^* &= W + \Delta^*,
\end{aligned}
\end{equation}
where $W$ denotes a selected subset of model parameters to be edited, $k_i$ is the hidden representation of the decisive token at the layer immediately preceding $W$, and $m_i$ is an idealized hidden representation aligned with the desired edit target. The second regularization term is designed to enforce orthogonality between $\Delta$ and other non-target knowledge.
It has a succinct solution:
\begin{equation}\label{eqa: memit solution}
    \Delta^*=(M_I-WK_I)K_I^{\top}(K_IK_I^{\top}+\lambda K_JK_J^{\top})^{-1},
\end{equation}
where $M_I,K_I,K_J$ represent the concatenations of all $m_i$, $k_i$, and $k_j$, respectively.

\section{Revisiting Specificity Evaluation}
\label{sec:specificity_revisit}
\subsection{A quick intuition of the common practice} \label{sec: privious practice}
This section will introduces how the specificity (preservation of non-target knowledge) is evaluated in the current literature with intuitive examples.
To the best of our knowledge, there are two major types of metrics for the specificity evaluation that are commonly used. 

The first one is \textbf{ground truth accuracy} \citep{memit}.  It evaluates specificity by posing a non-target factual question and measuring whether the edited model's answer matches human-annotated ground-truth string.
For instance, it samples a query $q=$``Who played Desmond Doss’ father in Hacksaw Ridge?'' and annotates the answer $y=$`` Hugo Weaving'' for it. And the specificity is measured by whether the answer tokens (i.e., Hugo Weaving) of
\begin{quote}
\textit{Who played Desmond Doss’ father in Hacksaw Ridge? Hugo Weaving}
\end{quote}
rank top-1 at their positions under the edited model. Sample-level (referred as \textbf{S-accuracy})or teacher-forcing token accuracy (referred as \textbf{T-accuracy}) is reported as the specificity. 
See Appendix \ref{app: example of metrics} for more detailed examples. See Appendix \ref{app: teacher forcing example} for what is ``teacher-forcing''.

The second one is \textbf{contrastive candidate comparison} \citep{memit}. This is another metric used for counterfactual datasets.  We first consider an example where the edit query is $q=$``Where does Watts Humphrey work?'', and the old knowledge is $y_{old}=$``Illinois Institute of Technology'', and the edit target is $y_{new}=$``Massachusetts Institute of Technology''. We then sample a neighborhood query $q'$ which has the same ground-truth answer as $q$. For example, both  Watts Humphrey and Mike work at Illinois Institute of Technology, and our $q'$ will be ``Where does Mike work?''. Specificity is assessed by comparing the likelihoods of
\begin{quote}
\textit{Where does Mike work? Massachusetts Institute of Technology}  
vs.  
\textit{Where does Mike work? Illinois Institute of Technology}.
\end{quote}

If the edited model assigns higher average token likelihood (or lower cross-entropy) to the latter sequence on the answer part, i.e., 
\begin{equation}
    P_{\theta^*}(y_{old}|x')>P_{\theta^*}(y_{new}|x'),
\end{equation}
the knowledge about Mike is considered not affected. We refer to this metric as \textbf{C-accuracy}. See Appendix \ref{app: example of metrics} for more detailed examples.

We call these existing specificity metrics as GT-based (GT is short for ground truth) metrics.

 \subsection{How can GT-based specificity evaluation be problematic}
This section will conceptually introduce the problems of GT-based specificity evaluation, and the empirical verification is in $\S$\ref{sec: verification of multi issues}.

The first straightforward issue is that there might be a gap between the dataset's knowledge and pre-edit model's knowledge. 
And beyond this surface issue, there are also more deeply hidden and inherent issues, which are hard to be addressed in GT-based evaluation.

The second problem is that the ground-truth accuracy metric is at odds with the open-ended nature of LLMs.
It requires the edited model’s output to align a definite predefined ground-truth token string, regardless of semantic equivalence (e.g., ``Massachusetts Institute of Technology'' is the right answer but ``MIT'' is not). It is built on two criteria: (1) There is a determined ground truth for the query $q$. (2) Any deviation from the given ground truth constitutes an error. This can be reasonable only if the following assumption holds: 
\begin{assumption}\label{ass: only one ground truth}
    For every test query $q\in \mathcal{Q}$, there exists a single answer $y_{GT}^q\in Y$ such that \\
    (i) $y_{GT}^q$ is the \textbf{unique} correct answer to $q$, and \\
    (ii) any deviation from $y_{GT}^q$ constitutes an error, regardless of the model’s pre-edit behavior.
\end{assumption}

However, Assumption \ref{ass: only one ground truth} does not always hold in open-ended LLM settings, where multiple semantically valid answers may exist and the model’s pre-edit behavior does not align with a specific token string.
Therefore, this metric is overly difficult, causing many methods to achieve uniformly low scores; as a result, their performances are flattened, and the metric lacks sufficient discriminative power.

At the other extreme, the contrastive candidate comparison metric is often overly permissive.
By evaluating whether the edited model prefers a specific reference answer over a small set of alternatives, these metrics only detect large, targeted shifts, such as the model adopting a particular new answer.
As long as the model does not collapse to that specific alternative, substantial distributional changes may go undetected.
Consequently, such metrics tend to saturate easily, assigning high specificity scores even when the model’s behavior has drifted significantly.

Moreover, contrastive candidate comparison metrics fail to isolate the model’s inherent, \textbf{query-independent biases}, leading to distorted specificity measurements.
For example, we denote (qualitative, informal illustration)
\begin{equation}
\log P(y \mid q)
=
\sum_{t}\log P(y_t \mid q,y_{i<t})
\end{equation}
The second term $y_{i<t}$ represents something like the inner-answer probability. For example, if ``Institute of Technology'' frequently appears after ``Massachusetts'' in the LLM's pretraining data, the metric’s outcome may be dominated by answer-internal fluency rather than query-conditioned factual grounding. This issue can affect both the contrastive candidate comparison metric and the teacher-forcing token-level accuracy metric.

And beyond the above issues, another limitation of existing specificity metrics lies in their \emph{discrete or ordinal nature}.
 Behavioral deviation is inherently continuous: edits may induce varying degrees of deviation in the model’s output distribution on unrelated queries.
However, existing metrics collapse this continuous deviation into coarse, discrete decisions (each sample only belongs to a binary label true or false), resulting in too hard or too easy test and thus systematic insensitivity.

\subsection{Theoretical perspective and new criterion}
\label{sec:method}
Given the limitations of GT-based specificity protocols, we need a new criterion to guide the design of specificity evaluation. 

We first formalize specificity as a counterfactual invariance property, and then analyze why commonly used GT–based specificity metrics fail to faithfully measure this property. Then, we present a concrete counterexample demonstrating that such metrics can report high specificity even when the model’s behavior with respect to unrelated queries has substantially changed.

\textbf{Specificity as behavioral deviation}. Let $f_\theta$ denote an LLM parameterized by $\theta$. A knowledge edit is an intervention that produces updated parameters $\theta^*=E(\theta)$, intended to modify the model’s behavior on a specific target fact.
Let $q$ be a query that is unrelated to the edited knowledge. Intuitively, an ideal specificity metric should depend only on the difference induced by the intervention, and does not assume access to any external ground-truth answer.
We formalize specificity as follows:
\begin{definition}\label{def: specificity}
Given a query $q$ whose knowledge needs to be preserved, the specificity of an edit $E$ is measured by the similarity between the model’s pre-edit and post-edit behaviors:
\begin{equation}\label{eqa: distance based specificity}
    \mathcal{L}_d(q)=d(f_\theta(q),f_{\theta^*}(q)), 
\end{equation}
where $d(\cdot,\cdot)$ is a deviation measure.

\end{definition}

\textbf{Theoretical perspective of GT-based specificity}. Most existing work (as introduced in $\S$\ref{sec: privious practice}) departs from Definition \ref{def: specificity} by introducing a predefined definite ground-truth answer $y_{GT}$ (the subscript $_{GT}$ is short for ground truth)
 for each test query $q$, and evaluates specificity via a loss function $l$:
 \begin{equation}\label{eqa: gt based specificity}
     \mathcal{L}_{GT} (q)= l(y_{GT},f_{\theta^*}(q))
 \end{equation}
Comparing Eq. \ref{eqa: gt based specificity} with Eq. \ref{eqa: distance based specificity}, the major difference lies in replacing a high-dimensional distribution $f_\theta(q)$ with a scalar $y_{GT}$. This constitutes an extreme form of compression that discards substantial information, which theoretically explains the issue described in Assumption \ref{ass: only one ground truth}. Moreover, since the model output is compressed into a single scalar, it is no longer suitable for distance-based, continuous-valued evaluation, and instead only supports binary true–false judgments. This provides a theoretical explanation for the discretization issue discussed at the end of the previous section. This is also analogous to the idea of knowledge distillation \citep{hinton2015distilling}, where continuous soft labels are more informative than one-hot labels.

\subsection{Empirical verification}\label{sec: verification of multi issues}
Previously, we have conceptually shown that GT-based specificity can potentially suffers from three problems: (1) The mismatch between the dataset's knowledge and the pre-edit model's knowledge. (2) Ignoring the open-ended nature of LLMs. (3) Query-independent biases. This subsection aims to investigate how often these problems can occur in practice. 

We conduct experiments using the widely adopted LLama3-8B-Instruct model\footnote{In the model editing literature, the widely used models are Llama3-8B-Instruct and GPT2-XL, and the widely used datasets are ZsRE and MCF. See $\S$\ref{sec: setup}.}. The dataset is the widely used ZsRE \citep{zsre}, and the results for another widely used dataset MCF \citep{rome} are in Appendix \ref{app: pre-edit mcf} (the findings are generally consistent). For each dataset, we sample  2,000 queries from their specificity evaluation sets. The pre-edit model first generates a response for each query, after which ``gpt-5-nano'' is used to judge whether the model’s response is consistent with the knowledge represented by the ground-truth answer provided in the dataset. 
We then split the consistent/inconsistent samples to form two new subsets. On the each subset, we evaluate the pre-edit model using existing specificity metrics
(the metrics suitable for ZsRE dataset is S-accuracy and T-accuracy, see Appendix \ref{app: example of metrics} for details of these metrics).

The results are shown in Table \ref{tab: preedit specificity}. From the table, several results merit attention. (1) First, for the widely used ZsRE dataset, only about $33\%$ of the data used for specificity evaluation contain knowledge that is consistent with the widely adopted Llama3 model. (2) Second, even within the filtered consistent subset, the probability that Llama3's response strictly matches the given definite ground truth (S-accuracy) is only about 0.5\%, highlighting the significant impact of the open-ended nature of LLMs. (3) Third, for the inconsistent subset, although the model does not possess the correct knowledge for the queried facts, the teacher-forcing token accuracy (T-accuracy) still reaches 37.6\%, demonstrating the non-negligible effect of query-independent biases. 

\begin{table}[]
    \centering
        \caption{The specificity performance of pre-edit llama3-8b-instruct on the ZsRE dataset. }
    \resizebox{0.99\columnwidth}{!}{
    \begin{tabular}{c c|c c}
    \hline
      \multicolumn{2}{c|}{ Consistent subset (33\%)}   &   \multicolumn{2}{c}{ Inconsistent subset (67\%)}\\
      \hline
        S-accuracy&T-accuracy & S-accuracy & T-accuracy\\
         \hline
         0.5 & 40.6 & 0.0 & 37.6 \\
         \hline
    \end{tabular}
    }

    \vspace{-10pt}
    \label{tab: preedit specificity}
\end{table}

\subsection{A simple and practical solution}
To address the limitations of GT-based evaluation, we propose to evaluate specificity without reference to external ground-truth answers, and to quantify it as a continuous measure of behavioral change. 

\textbf{Baseline metric: KL-divergence}.
Definition \ref{def: specificity} gives us a basic criterion about how to design the specificity metrics, and a straightforward solution for it is to use the KL-divergence to measure how much the post-edit behavior deviate from the pre-edit one. To avoid fluency biases, we use the logit of the query's last token to be the approximate query representation.\footnote{Although the last token’s logits are not equivalent to a representation of the entire query, they are a commonly used approximation in the LLM literature.} Then, the KL-divergence is calculated as follows:

\begin{equation}
    \mathcal{L}_{KL}(q)=D_{KL}(f_\theta(q)||f_{\theta^*}(q)).
\end{equation}

We do not claim novelty in proposing the KL-divergence metric; rather, we use it solely as a baseline to illustrate how a continuous, behavior-based metric should behave.

\textbf{Human-interpretable metric: approximated distance on support set}.
Considering that the absolute score of $\mathcal{L}_{KL}$ may not give a direct intuition for human, we additionally introduce a second metric. 
For distributions with heavy-tailed structure (such as LLMs' logits), agreement on the top-$k$ support implies bounded divergence in the region that most strongly influences generation. We denote $S_k(q,\theta)$ as the top-$k$ support of $f_\theta(q)$. We use the token overlap ration in the top-$k$ support as the second human-interpretable metric:
\begin{equation}
    \mathcal{L}_{top-k}(q)=\frac{S_k (q,\theta)\cap S_k (q,\theta^*)}{k}.
\end{equation}
Top-$k$ overlap is a linear transformation of the discrete L1 distance (i.e., the Hamming distance) computed on the logit-induced top-k support (please refer to Appendix \ref{app: L1 distance} for more detailed discussion). As such, it can be viewed as a form of sharpened distance, and thus is theoretically plausible supported by Eq. \ref{eqa: distance based specificity}.  Moreover, compared to $D_{KL}$, it preserves a percentage-based scale that eases human intuition.

\paragraph{Summary.} Conceptually, our Top-$k$ support based metric has the following advantages: (1) It is well-principled, as theoretically aligned with Definition \ref{def: specificity}. (2) By varying $k$, the metric provides a smooth, approximate interpolation toward a continuous formulation, allowing the test difficulty to be adjusted in a controlled manner. (3) It is free of query-independent biases, as it avoids teacher-forcing test. (4) It is easy to implement, as it does not rely on external annotations. Aside from the conceptual advantages, it also has strong empirical advantages, which is introduced in $\S$\ref{sec: experiments}.

\section{Experiments} \label{sec: experiments}
\subsection{Experimental setup}\label{sec: setup}
\textbf{Datasets}. We follow a recent work AlphaEdit \citep{alphaedit} to employ two widely used datasets: Multi-Counterfact (MCF) \citep{rome} and ZsRE \citep{zsre}, and do 2000 edits for each of the datasets.  Consistent with standard MEMIT practice, we perform massive editing (in which all knowledge items are edited jointly in a single batch), because sequential editing produces multiple intermediate and entangled $\Delta$ and $k$ states that can introduce unknown sources of noise, adding difficulties to isolate and observe the pure relationship between the specificity regularizer and the specificity metrics. See Appendix \ref{app: massive editing reason} for more detailed discussion.

\textbf{Models}. We employ three popular open-source models widely used in previous model editing literature: Llama-3-8B-Instruct, GPT2-XL, and Qwen2.5-7B-Instruct. 
Note that in this area we typically do not use the latest LLMs for experiments, because the most recent models may suffer from data leakage with respect to the evaluation datasets. Following the results of previous literature \citep{easyedit}, the decisive layers are the MLPs of the following layers: $[4,5,6,7,8]$ for Llama and Qwen, $[13,14,15,16,17]$ for GPT-XL. Due to space constraints, we primarily use Llama for our experiments, and the results for other models are provided in Appendix \ref{app: results with other models} (findings are generally consistent).

\begin{table*}[]
    \centering
    \caption{Sensitivity of different specificity metrics to the strength of the specificity regularizer. The dataset is MCF.
We vary the regularization coefficient $\lambda$ over two orders of magnitude and report the resulting regularizer values $\|\Delta k_j\|$ across layers.
While ground-truth-based specificity metrics show limited variation and weak alignment with the regularizer, our ground-truth-free metrics exhibit substantially higher sensitivity and perfect rank correlation.
}
    \resizebox{1.99\columnwidth}{!}{
    \begin{tabular}{c|c |c |c |c |c |c |c |c |c|c |c  }
    \hline
      \multirow{2}{*}{\diagbox{Methods}{Metrics}}   & \multicolumn{5}{c|}{  Specificity regularizer  $||\Delta k_j|| $ (mean)}& \multicolumn{2}{c|}{ GT-based specificity} & \multicolumn{4}{c}{ GT-free specificity  }\\
      \cline{2-12}
         & Layer 4 & Layer 5 & Layer 6 & Layer 7 & Layer 8 & S-acc $\uparrow$ & C-acc $\uparrow$& $D_{KL}$ $\downarrow$& Top-1 $\uparrow$&  Top-5 $\uparrow$& Top-10 $\uparrow$\\
         \hline
          pre-edit&0 &0 &0 & 0 &0&21.3 &89.3 &0 &100 &100 &100\\
         $\lambda=1.5\times10^6$&$2.9\times10^{-5}$&$4.7\times10^{-5}$&$8.6\times10^{-5}$&$1.7\times10^{-4}$&$5\times10^{-5}$&22.2 &89.7 & 0.015 &95.4&94.1&93.7\\
         $\lambda=1.5\times10^5$&0.001&0.002&0.003&0.006&0.017&23.8 &89.2&0.127&86.5&84.0&83.2\\
         $\lambda=1.5\times10^4$&0.023&0.033&0.052&0.096&0.263& 23.4&87.7&0.431&76.8&74.8&73.8\\
         $\lambda=1.5\times10^3$&0.068&0.100&0.163&0.297&0.650&23.1&87.2&0.539&74.3&72.3&71.2\\
         $\lambda=1.5\times10^2$&0.084&0.121&0.197&0.356&0.739&23.0&87.1&0.552&74.1&71.8&70.8\\
         \hline
         \\
         \hline
         Kendall’s $\tau$ correlation$^*$ $\uparrow$ & 1.00 & 1.00 & 1.00 & 1.00 & 1.00 &0.20 &0.87 & 1.00 &1.00 &1.00&1.00 \\
         Range & n/a & n/a & n/a & n/a & n/a & 0 $\sim$ 100& 0 $\sim$ 100 &n/a & 0 $\sim$ 100 & 0 $\sim$ 100 & 0 $\sim$ 100\\
         (Max-Min)/Range $\uparrow$& n/a & n/a & n/a & n/a & n/a & 2.5\% & 2.6\% & n/a &25.9\%& 28.2\% & 29.2\%\\
         std $\uparrow$& n/a & n/a& n/a& n/a & n/a & 0.9 & 1.2 &n/a &11.3 & 12.0 &12.4\\
         \hline
         \multicolumn{12}{l}{ $^*$: We report the absolute value of Kendall’s $\tau$ (i.e., ignoring the sign), and the mean across layers 4–8 is used for comparison.}\\
    \end{tabular}
    }
    
    \label{tab: correlation with specificity mcf llama}
\end{table*}

\begin{table*}[]
    \centering
        \caption{Sensitivity of different specificity metrics to the strength of the specificity regularizer. The dataset is ZsRE. ``T-acc'': teacher-forcing token accuracy. The ZsRE dataset is not a counterfactual dataset, so ``T-acc'' is used as the substitution of ``C-acc''.
}
    \resizebox{1.99\columnwidth}{!}{
    \begin{tabular}{c|c |c |c |c |c |c |c |c |c|c |c  }
    \hline
      \multirow{2}{*}{\diagbox{Methods}{Metrics}}   & \multicolumn{5}{c|}{  Specificity regularizer  $||\Delta k_j|| $ (mean)}& \multicolumn{2}{c|}{ GT-based specificity} & \multicolumn{4}{c}{ GT-free specificity  }\\
      \cline{2-12}
         & Layer 4 & Layer 5 & Layer 6 & Layer 7 & Layer 8 & S-acc $\uparrow$ & T-acc $\uparrow$& $D_{KL}$ $\downarrow$& Top-1 $\uparrow$&  Top-5 $\uparrow$& Top-10 $\uparrow$\\
         \hline
          pre-edit&0 &0 &0 & 0 &0&0.2 &38.5 &0 &100 &100 &100\\
         $\lambda=1.5\times10^6$&$4.8\times10^{-5}$&$7.5\times10^{-5}$&$1.4\times10^{-4}$&$2.9\times10^{-4}$&$9.4\times10^{-4}$&0.6 &39.4 & 0.046 &82.6&90.4&91.8\\
         $\lambda=1.5\times10^5$&0.001&0.002&0.003&0.006&0.017&3.6 &42.2&0.448&56.0&72.7&76.5\\
         $\lambda=1.5\times10^4$&0.013&0.019&0.032&0.063&0.185&5.2&43.4&0.659&43.6&65.4&70.3\\
         $\lambda=1.5\times10^3$&0.057&0.094&0.167&0.330&0.880&4.6&43.1&0.761&38.4&60.1&66.8\\
         $\lambda=1.5\times10^2$&0.402&0.380&0.366&0.490&0.941&4.8&42.9&1.032&29.4&52.7&59.7\\
         \hline
         \\
         \hline
         Kendall’s $\tau$ correlation $\uparrow$ &1.00&1.00&1.00&1.00&1.00&0.73&0.60&1.00&1.00&1.00&1.00 \\
          Range & n/a & n/a & n/a & n/a & n/a & 0 $\sim$ 100& 0 $\sim$ 100 &n/a & 0 $\sim$ 100 & 0 $\sim$ 100 & 0 $\sim$ 100\\
         (Max-Min)/Range $\uparrow$& n/a & n/a & n/a & n/a & n/a &5.0\% & 4.9\% & n/a &70.6\%& 47.3\% & 40.3\%\\
         std $\uparrow$ & n/a & n/a& n/a& n/a & n/a & 2.2 & 2.1 &n/a &27.5 & 18.3 &15.5\\
         \hline
    \end{tabular}
    }

    \label{tab: correlation with specificity zsre llama}
\end{table*}

\subsection{Sensitivity and alignment}
\label{sec:sensitivity}

We next study how different specificity metrics respond to changes in the strength of the specificity regularizer.
Recall the editing objective in Eq. \ref{eq: memit} (i.e., standard MEMIT \citep{memit}), where the hyperparameter $\lambda$ controls the trade-off between editing efficacy and specificity preservation.
Following common practice in prior work, $\lambda$ is typically set to $1.5 \times 10^4$.
To probe metric sensitivity, we vary $\lambda$ by two orders of magnitude in both directions, ranging from $1.5 \times 10^2$ to $1.5 \times 10^6$, and examine how different specificity metrics change as the magnitude of the specificity regularizer varies.

Table \ref{tab: correlation with specificity mcf llama} and \ref{tab: correlation with specificity zsre llama} report the resulting regularizer values $\|\Delta k_j\| $ across layers, together with different specificity metrics.
As expected, the regularizer values change substantially with $\lambda$, indicating progressively stronger constraints. 

\paragraph{Pre-edit behavior.}
We first examine the pre-edit model as a reference point.
Notably, the ground-truth accuracy (S-accuracy) on non-target queries is extremely low (around 21\%), implying a false-negative rate close to 80\%.
This observation corroborates our earlier analysis: treating ground-truth answer as the evaluation target ignores the open-ended nature of LLM outputs, and penalizes semantically valid but surface-divergent generations. Similarly, the contrastive comparison metric (C-accuracy) does not achieve 100\% even for the pre-edit model.
This suggests that the metric is influenced by query-independent biases, such as answer-internal language priors, rather than purely reflecting entity-specific knowledge.
Both effects are further amplified by implicitly assuming that the pre-edit model knows the test knowledge. Interestingly, under ground-truth-based specificity metrics, the pre-edit model does not always achieve the highest score, a phenomenon also happens in prior studies \citep{alphaedit,residutaledit}.
This further highlights the difficulty of using ground-truth-based metrics to reliably compare specificity across methods.

\paragraph{Sensitivity under varying regularization strength.}
We now turn to the core question: how sensitive are different specificity metrics to changes in the strength of specificity regularization? From Table \ref{tab: correlation with specificity mcf llama}, we observe that the two GT-based specificity metrics exhibit only marginal changes with respect to $\lambda$. Is it because the regularizer $\|\Delta k_j\| $ itself is insensitive to $\lambda$? The answer is No. Unlike the specificity metrics, the regularizer values $\|\Delta k_j\| $ generally vary smoothly and monotonically with $\lambda$ (so as $D_{KL}$), showing that $\lambda$ does have a considerable effect on specificity.
However, the GT-based scores remain largely saturated across a wide range of regularization strengths, making it difficult to distinguish edits with substantially different impacts on unrelated knowledge.

This lack of sensitivity is problematic in practice.
It may lead to misleading conclusions, for example, suggesting that a method achieves large gains in editing efficacy with ``almost no'' specificity loss, when in fact the underlying model behavior has changed significantly. One real-world example is AlphaEdit, which discards the specificity regularizer by constructing a null space of non-target knowledge \citep{alphaedit}. However, this null-space construction is only approximate and cannot fully preserve non-target knowledge. As a result, AlphaEdit achieves specificity comparable to MEMIT under GT-based  metrics, yet exhibits a non-ignorable gap from MEMIT in terms of $D_{KL}$, indicating incomplete protection of non-target knowledge. In contrast, our top-k metrics, consistent with $D_{KL}$, are able to clearly capture this discrepancy.

In contrast, our GT-free specificity metrics respond consistently and smoothly to changes in $\lambda$ and $||\Delta k_j|| $.
All the top-$k$ overlap metrics exhibit clear and monotonic trends as the regularization strength decreases, showing behavior similar to that of $D_{KL}$.
This higher resolution enables more faithful discrimination between edits with different degrees of specificity preservation.

To quantitatively assess sensitivity, we additionally report two scores: coverage proportion ($\frac{Max-Min}{Range}$), which measures the range of variation under different experimental settings, and standard deviation. From Table \ref{tab: correlation with specificity mcf llama}, We observe that GT-based specificity exhibits very small coverage proportion and standard deviation, even when the strength of the specificity regularizer varies substantially.

\paragraph{Metric alignment with specificity regularization.}
Beyond sensitivity, another important criterion is ranking consistency. In general, as the strength of the specificity regularizer increases, specificity is expected to improve accordingly. Using the average $||\Delta k_j|| $ of the edited layers as a reference, we report the Kendall’s $\tau$ correlation for different metrics.
We choose Kendall’s $\tau$ over Pearson or Spearman correlation as it is non-parametric, robust to non-linear scaling, and more suitable for situations with very small sample sizes. Table \ref{tab: correlation with specificity mcf llama} shows that GT-free metrics exhibit perfect rank correlation with the specificity regularizer strength.
By contrast, GT-based metrics show  weak correlation.

\begin{table*}[t]
    \centering
       \caption{Benchmarking different editing methods. The model is Llama3-8b-instruct.
}
    \resizebox{1.99\columnwidth}{!}{
    \begin{tabular}{c| c |c |c |c |c |c |c |c  }
    \hline
     \multicolumn{9}{c}{ The MCF dataset}
    \\
    \hline
    
      \multicolumn{1}{c|}{\multirow{2}{*}{\diagbox{Methods}{Metrics}}}   & \multicolumn{2}{c|}{ Target knowledge}& \multicolumn{2}{c|}{ GT-based specificity} & \multicolumn{4}{c}{ GT-free specificity  }\\
      \cline{2-9}
         \multicolumn{1}{c|}{} & Efficacy $\uparrow$& Generalization $\uparrow$& S-acc $\uparrow$ & C-acc $\uparrow$& $D_{KL}$ $\downarrow$& Top-1 $\uparrow$&  Top-5 $\uparrow$& Top-10 $\uparrow$\\
         \hline
          \multicolumn{1}{c|}{pre-edit}&7.6& 10.5&21.3 &89.3&0 &100 & 100 &100\\
          \multicolumn{1}{c|}{WISE} & 18.0 & 11.8 & 20.9 &83.4 & 0.09 & 97.7 & 97.5 & 97.4\\
          \multicolumn{1}{c|}{RLedit} & 57.8 &51.5 &10.6 & 49.2 & 5.62 & 2.57 &8.6 &9.8\\
          \multicolumn{1}{c|}{MEMIT}& 97.4 & 80.3& 23.4& 87.7 & 0.43& 76.8&74.8&73.8\\
          \multicolumn{1}{c|}{Adaedit }& 99.0 & 91.6 &21.8&84.1& 1.09 & 62.4 & 60.6 &59.7\\
          \multicolumn{1}{c|}{Alphaedit} & 95.6& 77.2& 24.1&87.2& 0.63&71.2&68.9&67.9\\
          \multicolumn{1}{c|}{EMMET} & 99.8& 89.7&23.2& 87.3&0.54&74.4&72.3&71.3\\
          \multicolumn{1}{c|}{NAMET} & 97.4&80.4&23.4&87.7&0.43& 76.8&74.8&73.8\\
          \multicolumn{1}{c|}{PMET} & 99.2& 92.3 & 21.9&83.9 & 1.15 & 61.2 &59.8 &58.8\\ 
          \multicolumn{1}{c|}{PRUNE} & 97.5 & 80.2 & 23.4 &87.8 &0.43 & 76.8 & 74.8 &73.8\\
          \multicolumn{1}{c|}{RECT }& 96.8 & 79.2 &23.5 &87.9 & 0.42 &76.9 & 75.1 &74.1\\
          \hline
\multicolumn{9}{c}{}
\\
\hline
     \multicolumn{9}{c}{ The ZsRE dataset}
    \\
    \hline
               \multicolumn{1}{c|}{\multirow{2}{*}{\diagbox{Methods}}{Metrics}}   & \multicolumn{2}{c|}{ Target knowledge}& \multicolumn{2}{c|}{ GT-based specificity} & \multicolumn{4}{c}{ GT-free specificity  }\\
      \cline{2-9}
         \multicolumn{1}{c|}{}& Efficacy $\uparrow$& Generalization $\uparrow$& S-acc $\uparrow$ & T-acc $\uparrow$& $D_{KL}$ $\downarrow$& Top-1 $\uparrow$&  Top-5 $\uparrow$& Top-10 $\uparrow$\\
         \hline
          pre-edit& 38.1 & 37.6& 0.2 & 38.6 &0 &100 &100 &100\\
          WISE & 40.6 & 2.0 & 0.1 & 39.3 & 1.75 & 53.1 & 57.8 & 57.5\\
          RLedit & 71.0 &69.2 & 5.0 & 36.7 & 3.51 & 1.65 & 8.5 &15.1\\
          MEMIT& 92.3 & 88.4 &5.1 &43.3 & 0.59 &46.0 & 66.6 &71.6\\
          Adaedit & 92.7 & 90.3 & 10.0 &48.0 & 2.19 &13.8 &34.5 &43.6\\
          Alphaedit & 90.8 & 86.9 & 6.8 &45.3 & 0.92 & 32.6 &56.9 & 63.8\\
          EMMET & 97.0 &92.1 &4.3 &42.7 &0.69 &42.7 & 63.8 & 68.7\\
          NAMET & 92.5 & 88.5 & 5.6 &43.7 &0.61 &44.1 &66.1 &71.2\\
          PMET & 93.2 & 90.6 &9.9 &47.8 & 2.29 &13.1 &33.6 &42.2\\ 
          PRUNE & 92.5 & 88.4 & 5.2& 43.3 & 0.59 &46.2 &66.9 &71.7\\
          RECT& 91.8 & 88.0 &5.1 &43.3 & 0.58 &46.6 &67.2 &71.8\\
          
             \hline
\multicolumn{9}{c}{}
\\
\hline
     \multicolumn{9}{c}{ The rank stability (Kendall's $\tau$) of different methods across datasets}
    \\
    \hline
               \multirow{2}{*}{Metrics}   & \multicolumn{2}{c|}{ Target knowledge}& \multicolumn{2}{c|}{ GT-based specificity} & \multicolumn{4}{c}{ GT-free specificity  }\\
      \cline{2-9}
         \multicolumn{1}{c|}{}& Efficacy $\uparrow$& Generalization $\uparrow$& S-acc $\uparrow$ & T-acc $\uparrow$& $D_{KL}$ $\uparrow$& Top-1 $\uparrow$&  Top-5 $\uparrow$& Top-10 $\uparrow$\\
    \hline    
    & 0.98 &0.88&0.28&-0.09 &0.75&0.97&0.79&0.75\\
    \hline
    Average &  \multicolumn{2}{c|}{0.93}& \multicolumn{2}{c|}{0.10} & \multicolumn{4}{c}{ 0.82 }\\
    \hline

    \end{tabular}
    }
 
    \label{tab: benchmark mcf llama}
\end{table*}

\subsection{Benchmarking different editing methods}

Having established that ground-truth-free specificity metrics are more sensitive and better aligned with specificity regularization, we now investigate whether they enable more meaningful comparison across different knowledge editing methods.
Specifically, we ask whether different editing approaches, designed with varying mechanisms and inductive biases, can be reliably distinguished in terms of their impact on non-target knowledge.

\textbf{Editing methods}. We employ ten representative model editing methods published in recent years,  including MEMIT \citep{memit}, WISE \citep{wise},  RECT \citep{rect}, EMMET\citep{gupta2024unified}, PMET\citep{pmet}, PRUNE \citep{prune}, Adaedit \citep{adaedit}, Alphaedit \citep{alphaedit}, NAMET \citep{namet}, RLedit \citep{RLEDIT}. Except for the specificity metrics, we follow Alplaedit \citep{alphaedit} to use Efficacy and Generalization as the metrics to measure the ratio of successful editing of target knowledge. It is worth noting that ZsRE is not a counterfactual dataset and therefore does not include the ``C-accuracy'' metric. In prior literature, the commonly used substitute is ``T-accuracy''. The similarity to ``C-accuracy'' lies in that both compute the average success rate at each position under a teacher-forcing setting.

\textbf{Results}. The results on the MCF and ZsRE datasets are in Table \ref{tab: benchmark mcf llama}. In general, a method’s capability in a particular aspect is intrinsic. Therefore, although the ranking of different methods may vary somewhat across datasets, it should remain largely consistent overall. We use Kendall’s $\tau$ to quantify the consistency of method rankings across different datasets. The metrics related to target knowledge injection exhibit high Kendall’s $\tau$ across the two datasets, which to some extent indicates that the evaluated methods possess a certain level of generalizability across these datasets. However, the GT-based specificity metrics show very low Kendall’s $\tau$, suggesting that the rankings of different methods across datasets are almost random. This is not because the locality behavior of these methods is entirely random across different datasets, as evidenced by the high Kendall’s $\tau$ of $D_{KL}$.
This makes it difficult for these results to provide constructive insights into the relative strengths and weaknesses of different methods. 
In contrast, the GT-free specificity metrics achieve Kendall’s $\tau$ values that are comparable to those of the target-knowledge-related metrics to some extent, thereby validating the advantages of GT-free specificity metrics.

\section{Limitations}

Designing a perfect metric is inherently difficult, and progress is necessarily incremental.
While our ground-truth-free metrics address several problems of existing evaluations, they also have their own limitations.
First, we use the output logits at the final token position as the representation of the query.
Although this choice is simple and widely applicable, it does not capture all aspects of generation, such as longer-range dependencies or alternative valid continuations.
Nevertheless, we think it is a good trade-off between completeness and practical usability.
Second, we use the strength of the specificity regularizer as a reference signal when analyzing metric sensitivity and alignment.
Regularizer strength does not uniquely determine final specificity, but it is intended to control representation drift and is therefore reasonably correlated with locality in practice.
Overall, our metrics are not intended to be definitive, but rather to provide a more informative and sensitive alternative to commonly used specificity evaluations under current practice.

\section{Impact Statements}
This paper presents work whose goal is to advance the field of machine learning. There are many potential societal consequences of our work, none of which we feel must be specifically highlighted here.

\bibliography{example_paper}
\bibliographystyle{icml2026}

\clearpage
\appendix

\section{Related Works}

\textbf{RAG-style approaches for inference-time knowledge editing}.
A line of work draws inspiration from Retrieval-Augmented Generation (RAG), where an external knowledge base is maintained and queried during generation to provide updated information \cite{hartvigsen2023aging, zheng2023can, zhang2024instructedit, jiang2024learning,zhaoincontext}. While effective in some settings, these methods inherit the limitations of RAG, including reliance on retrieval quality, context-length constraints, and increased system latency and complexity. As they do not modify model parameters, we consider them RAG-based systems rather than genuine model editing methods, and thus exclude them from our scope.

\textbf{Augmenting models with extra memory or hypernetworks.}
Another stream augments language models with auxiliary memory modules or hypernetworks to store new knowledge. Representative methods include MEND \citep{mitchellfast}, KE \citep{de2021editing}, and RLEdit \citep{RLEDIT}, which often employ low-rank updates for efficiency. However, hypernetwork-based approaches typically generalize poorly, requiring retraining or finetuning for each new fact. Related methods that store knowledge in auxiliary parameters or networks—such as T-Patcher \citep{huangtransformer}, SERAC \citep{mitchell2022memory}, Grace \citep{hartvigsen2023aging}, WISE \citep{wise}, and KDE \citep{kde}, suffer from scalability issues, as knowledge accumulates without removing outdated information, and generally achieve lower editing precision than exact parameter-editing methods.

\textbf{Precise model (parameter) editing.}
Recent work has increasingly focused on precise parameter editing, typically following a \emph{locate-then-edit} paradigm: identifying decisive tokens and layers and directly modifying the corresponding hidden states or parameters to substitute target knowledge. Early methods such as ROME and MEMIT \citep{rome,memit} introduced causal tracing to locate effective edit positions. Subsequent approaches, including RECT \citep{rect}, EMMET \citep{gupta2024unified}, PMET \citep{pmet}, PRUNE \citep{prune}, AdaEdit \citep{adaedit}, AlphaEdit \citep{alphaedit}, NAMET \citep{namet}, MEMIT-LTI \citep{overfit},AnyEdit \citep{anyedit}, and so on \citep{neuraledit,explainedit,diffusionedit}, extend this framework with improved optimization or regularization strategies. Owing to their parameter efficiency and strong empirical performance, these methods have become the dominant paradigm, with recent extensions to more complex tasks such as multi-hop reasoning \citep{chainedit,overfit}.

\textbf{Evaluation and and other aspects}. Currently, most research on evaluation and benchmarking primarily focuses on the target knowledge aspect. UniEdit \citep{uniedit} proposes a unified, open-domain benchmark that standardizes the evaluation of knowledge editing across diverse domains. WILD \citep{wild} argues that the editing efficacy should be evaluated  with LLM as a judge. ThinkEval \citep{thinkeval} introduces a graph-based evaluation framework. \cite{editsafe} warn that model editing may introduce safety risks. \cite{detectedit} propose a method for detecting whether a given fact has been edited.

\section{The choice of massive editing}\label{app: massive editing reason}

We follow the standard MEMIT to choose the massive editing setting, because it is suited for controlled, variable-isolated analysis. 
For example, $K_J$ in Eq. \ref{eqa: memit solution} is independent of the editing methods and should be regarded as a data pre-processing output that can be shared across all methods. However, in sequential editing, each edit step in practice induces a drift in $K_J$. While to reduce computational cost, it is common to use the value computed in the first step. Consequently, sequential editing is ill-suited for our controlled and fine-grained analysis, as it introduces confounding factors that hinder isolating the effects of specificity regularizers and specificity metrics. In addition, under the massive editing setting, all methods can share the $M_I-WK_I$ term in Eq. \ref{eqa: memit solution}, which makes it particularly well suited for controlled, variable-isolated analysis.

\section{More detailed examples of GT-based specificity}\label{app: example of metrics}
Generally, there are three commonly used specificity metrics, which we denote as Accuracy, Comparison, T-accuracy.

We still use the following example (the factual knowledge that needs to be preserved is that Mike works at Illinois Institute of Technology):
\begin{quote}
\textit{Where does Mike work? Massachusetts Institute of Technology}  
vs.  
\textit{Where does Mike work? Illinois Institute of Technology}.
\end{quote}

\textbf{S-accuracy}. S-accuracy is short for ``sample accuracy''.
We use ``Where does Mike work?'' as the input of the post-edit LLM. If the LLM generates exactly the token string `` Illinois Institute of Technology'', the answer is considered correct, otherwise it is considered incorrect.

\textbf{C-accuracy}. We use ``Where does Mike work? Massachusetts Institute of Technology'' as the input of the post-edit LLM. For each token at the positions of ``Massachusetts Institute of Technology'', we will have a probability for generating the token. We multiply the probabilities of each token in the answer section to obtain the probability of this answer occurring. We denote it as $P_1$. In the same way, we use ``Where does Mike work? Illinois Institute of Technology'' as the input and get another probability $P_2$. If $P_1<P_2$, we consider it as a success (as the factual knowledge does not change). And the ratio of success in the dataset is the score for the Comparison metric.

\textbf{T-accuracy}. T-accuracy is short for ``teacher-forcing token accuracy'' (see Appendix \ref{app: teacher forcing example} for the definition of ``teacher-forcing''). In non-counterfact datasets, we only have ``Where does Mike work? Illinois Institute of Technology'' and do not have a paired example like ``Where does Mike work? Massachusetts Institute of Technology''. Therefore, the ``C-accuracy'' metric cannot be computed, and ``T-accuracy'' is used as a substitute. T-accuracy is defined as follows: given the input ``Where does Mike work? Illinois Institute of Technology'', we examine whether each token of “Illinois Institute of Technology” is the top-1 probability token at its corresponding position. If so, the token is marked as correct. Finally, the proportion of correct tokens in the dataset is reported as the T-accuracy.
Compared to T-accuracy, S-accuracy is more strict because it will count a false sample even if only one token is incorrect.  

Compared to \emph{S-accuracy}, both \emph{T-accuracy} and \emph{C-accuracy} are easier metrics. At the same time, they are both affected by query-independent biases.
In the MCF dataset, we generally use \emph{C-accuracy}, whereas in ZsRE, we typically use \emph{T-accuracy}. While \emph{S-accuracy} can be applied to both datasets, it is rarely used. Perhaps it's because it's too difficult, and all the methods have very low scores.

\section{Preliminaries of teacher-forcing generation} \label{app: teacher forcing example}
\begin{figure}[t]
\centering
    \includegraphics[width=0.99\columnwidth]{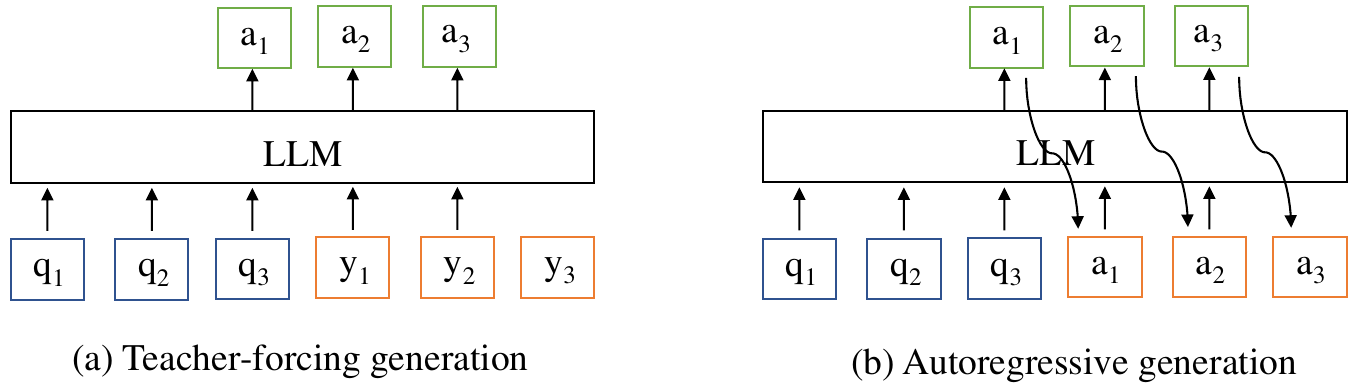}
    \caption{An example of teacher-forcing generation.}
    \label{fig: teacher forcing}
\end{figure}

LLMs typically generate answers in an autoregressive manner: the initial input is the query, and at each step, the newly generated token is appended to the query as part of the input to generate the next token. However, in certain evaluation settings, teacher-forcing generation is also used. Figure \ref{fig: teacher forcing} is an example.
Teacher forcing refers to concatenating the query ($q_1,q_2,q_3$) with a pre-specified ground-truth answer ($y_1,y_2,y_3$) as the input, and then performing a single forward pass to generate all answer tokens ($a_1,a_2,a_3$) at once. Under teacher forcing, the generation of subsequent tokens does not depend on previously generated tokens; instead, each token is predicted independently. 

\section{The pre-edit model's specificity on the MCF dataset}\label{app: pre-edit mcf}
Corresponding to Table \ref{tab: preedit specificity}, we report the results on the MCF dataset in Table \ref{tab: preedit specificity mcf}. On this dataset, the two common specificity metrics are S-accuracy and C-accuracy (see Appendix \ref{app: example of metrics}).

\section{More results}\label{app: results with other models}
Corresponding to Table \ref{tab: correlation with specificity mcf llama} and Table \ref{tab: correlation with specificity zsre llama}, the results of GPT2-XL and Qwen are in Tables \ref{tab: correlation with specificity mcf gpt2-xl}, \ref{tab: correlation with specificity zsre gpt2-xl}, \ref{tab: correlation with specificity mcf qwen}, and \ref{tab: correlation with specificity zsre qwen}.  

Also, corresponding to Table \ref{tab: benchmark mcf llama}, the same experiments for GPT2-XL and Qwen are in Tables \ref{tab: benchmark gpt2-xl} and \ref{tab: benchmark mcf qwen}.

\section{Top-$k$ as distance}\label{app: L1 distance}
\paragraph{Set-based interpretation.}
Let $A = \mathrm{Top}\text{-}k(f_{\theta}(q))$ and 
$B = \mathrm{Top}\text{-}k(f_{\theta^*}(q))$ denote the sets of top-$k$ tokens
before and after editing, respectively.
The top-$k$ overlap metric used in this work is defined as
\begin{equation}
\mathrm{Overlap}_k(A,B) = \frac{|A \cap B|}{k}.
\end{equation}
This metric admits a simple set-based interpretation.
Let $\mathbf 1_A, \mathbf 1_B \in \{0,1\}^{|\mathcal V|}$ denote the indicator vectors of
the two sets.
Then the $\ell_1$ distance between these indicator vectors satisfies
\begin{equation}
\|\mathbf 1_A - \mathbf 1_B\|_1 = 2\big(k - |A \cap B|\big).
\end{equation}
Consequently, the top-$k$ overlap can be equivalently expressed as
\begin{equation}
\mathrm{Overlap}_k(A,B)
= 1 - \frac{\|\mathbf 1_A - \mathbf 1_B\|_1}{2k}.
\end{equation}
Therefore, top-$k$ overlap is a linear transformation of the $\ell_1$ distance
between the indicator vectors of the top-$k$ supports.
From this perspective, it defines a normalized set-based distance over the dominant
support of the output distribution, measuring how much the identity of the most
preferred tokens changes under editing, without relying on probability magnitudes
or token ordering.

\begin{table}[]
    \centering
    \resizebox{0.99\columnwidth}{!}{
    \begin{tabular}{c c|c c}
    \hline
      \multicolumn{2}{c|}{ Consistent subset (51\%)}   &   \multicolumn{2}{c}{ Inconsistent subset (49\%)}\\
      \hline
         S-accuracy&C-accuracy& S-accuracy & C-accuracy\\
         \hline
         35.7 & 97.6 & 0.0 & 81.2 \\
         \hline
    \end{tabular}
    }
    \caption{The specificity performance of pre-edit llama3-8b-instruct on the MCF dataset. }
    \label{tab: preedit specificity mcf}
\end{table}

\begin{table*}[]
    \centering
    
    \resizebox{1.99\columnwidth}{!}{
    
    \begin{tabular}{c|c |c |c |c |c |c|c |c |c |c|c |c  }
    \hline
      \multirow{2}{*}{\diagbox{Methods}{Metrics}}   & \multicolumn{6}{c|}{  Specificity regularizer  $||\Delta k_j|| $ (mean)}& \multicolumn{2}{c|}{ GT-based specificity} & \multicolumn{4}{c}{ GT-free specificity  }\\
      \cline{2-13}
         & Layer 13 & Layer 14 & Layer 15 & Layer 16 & Layer 17&Average & S-acc $\uparrow$ & C-acc $\uparrow$& $D_{KL}$ $\downarrow$& Top-1 $\uparrow$&  Top-5 $\uparrow$& Top-10 $\uparrow$\\
         \hline
          pre-edit&0&0&0&0&0&0&10.5&77.8&0&100&100&100\\
         $\lambda=1.5\times10^6$&0.007&0.010&0.019&0.041&0.148&0.045&10.5&77.8&0.012&95.6&94.1&93.9\\
         $\lambda=1.5\times10^5$&0.333&0.467&0.810&1.588&4.849&1.610&13.1&77.7&0.314&77.6&74.6&73.4\\
         $\lambda=1.5\times10^4$&7.746&10.255&15.642&27.580&74.707&27.186&12.0&73.3&0.807&63.5&60.8&59.4\\
         $\lambda=1.5\times10^3$&55.127&68.257&82.402&113.974&224.200&108.792&9.4&69.8&1.020&59.2&55.9&54.2\\
         $\lambda=1.5\times10^2$&134.525&145.932&133.138&145.348&236.470&159.082&7.9&68.4&1.116&57.9&53.9&51.9\\
         \hline
         \\
         \hline
         Kendall’s $\tau$ correlation $\uparrow$ &1.00&1.00&1.00&1.00&1.00&n/a&0.41&0.97&1.00&1.00&1.00&1.00\\
         Range & n/a& n/a& n/a& n/a& n/a& n/a&0 $\sim$ 100 &0 $\sim$ 100 &n/a &0 $\sim$ 100&0 $\sim$ 100&0 $\sim$ 100
         \\
         (Max-Min)/Range $\uparrow$&  n/a& n/a& n/a& n/a& n/a& n/a&0.052 &0.094 &n/a &0.421&0.461&0.481 \\
         Standard deviation $\uparrow$ &  n/a& n/a& n/a& n/a& n/a& n/a&1.7 &3.9 &n/a &17.0&18.2&18.9 \\
         \hline

    \end{tabular}
    }
    \caption{Sensitivity of different specificity metrics to the strength of the specificity regularizer. The model is GPT2-XL, and the dataset is MCF.
}
    \label{tab: correlation with specificity mcf gpt2-xl}
\end{table*}

\begin{table*}[]
    \centering
    
    \resizebox{1.99\columnwidth}{!}{
    
    \begin{tabular}{c|c |c |c |c |c |c |c |c |c|c|c |c  }
    \hline
      \multirow{2}{*}{\diagbox{Methods}{Metrics}}   & \multicolumn{6}{c|}{  Specificity regularizer  $||\Delta k_j|| $ (mean)}& \multicolumn{2}{c|}{ GT-based specificity} & \multicolumn{4}{c}{ GT-free specificity  }\\
      \cline{2-13}
         & Layer 13 & Layer 14 & Layer 15 & Layer 16 & Layer 17&Average & S-acc $\uparrow$ & T-acc $\uparrow$& $D_{KL}$ $\downarrow$& Top-1 $\uparrow$&  Top-5 $\uparrow$& Top-10 $\uparrow$\\
         \hline
          pre-edit&0&0&0&0&0&0&0&24.9&0&100&100&100\\
         $\lambda=1.5\times10^6$&0.013&0.018&0.035&0.077&0.272&0.083&0&25.3&0.008&99.4&93.9&95.4\\
         $\lambda=1.5\times10^5$&0.466&0.641&1.026&1.849&5.240&1.844&0&26.2&0.344&90.0&64.4&67.9\\
         $\lambda=1.5\times10^4$&6.222&8.194&11.776&19.809&50.781&19.357&0.4&26.5&0.904&70.6&47.1&49.5\\
         $\lambda=1.5\times10^3$&33.727&43.884&57.524&90.436&211.570&87.428&0.1&25.7&1.052&66.6&43.3&46.0\\
         $\lambda=1.5\times10^2$&155.794&182.169&193.565&214.325&296.686&208.508&0.4&23.8&1.349&56.5&33.9&36.6\\
         \hline
         \\
         \hline
         Kendall’s $\tau$ correlation $\uparrow$ &1.00 &1.00 &1.00&1.00 &1.00& n/a &0.70 &0.07 &1.00 & 1.00&1.00&1.00 \\
         Range & n/a& n/a& n/a& n/a& n/a& n/a&0 $\sim$ 100 &0 $\sim$ 100 &n/a &0 $\sim$ 100&0 $\sim$ 100&0 $\sim$ 100
         \\
         (Max-Min)/Range $\uparrow$&  n/a& n/a& n/a& n/a& n/a& n/a&0.004 &0.027 &n/a &0.436&0.661&0.634 \\
         Standard deviation $\uparrow$ &  n/a& n/a& n/a& n/a& n/a& n/a&0.2 &0.9 &n/a &16.8&25.2&24.4 \\
         \hline
    \end{tabular}
    }
    \caption{Sensitivity of different specificity metrics to the strength of the specificity regularizer. The model is GPT2-XL, and the dataset is ZsRE. 
}
    \label{tab: correlation with specificity zsre gpt2-xl}
\end{table*}

\begin{table*}[]
    \centering
    \resizebox{1.99\columnwidth}{!}{
    \begin{tabular}{c|c |c |c |c |c |c|c |c |c |c|c |c  }
    \hline
      \multirow{2}{*}{\diagbox{Methods}{Metrics}}   & \multicolumn{6}{c|}{  Specificity regularizer  $||\Delta k_j|| $ (mean)}& \multicolumn{2}{c|}{ GT-based specificity} & \multicolumn{4}{c}{ GT-free specificity  }\\
      \cline{2-13}
         & Layer 4 & Layer 5 & Layer 6 & Layer 7 & Layer 8&Average & S-acc $\uparrow$ & C-acc $\uparrow$& $D_{KL}$ $\downarrow$& Top-1 $\uparrow$&  Top-5 $\uparrow$& Top-10 $\uparrow$\\
         \hline
          pre-edit&0&0&0&0&0&0&14.8&85.8&0&100&100&100\\
         $\lambda=1.5\times10^6$&0.041&0.069&0.111&0.158&0.553&0.186&15.4&85.5&0.029&92.4&92.1&92.0\\
         $\lambda=1.5\times10^5$&1.602&2.333&4.419&6.920&22.792&7.613&13.9&81.1&0.398&74.5&74.7&74.4\\
         $\lambda=1.5\times10^4$&16.708&22.475&31.109&53.951&90.035&42.856&10.1&73.8&0.999&61.2&61.0&60.0\\
         $\lambda=1.5\times10^3$&49.398&50.996&45.518&74.393&79.536&59.968&7.8&69.7&1.302&55.4&55.1&53.6\\
         $\lambda=1.5\times10^2$&67.375&53.672&46.085&76.281&78.480&64.379&7.0&68.5&1.391&53.8&53.6&52.2\\
         \hline
         \\
         \hline
         Kendall’s $\tau$ correlation $\uparrow$ &1.00&1.00&1.00&1.00&0.60&n/a&0.87&1.00&1.00&1.00&1.00&1.00\\
         Range & n/a& n/a& n/a& n/a& n/a& n/a&0 $\sim$ 100 &0 $\sim$ 100 &n/a &0 $\sim$ 100&0 $\sim$ 100&0 $\sim$ 100
         \\
         (Max-Min)/Range $\uparrow$&  n/a& n/a& n/a& n/a& n/a& n/a&0.084 &0.173 &n/a &0.462&0.464&0.478 \\
         Standard deviation $\uparrow$ &  n/a& n/a& n/a& n/a& n/a& n/a&3.7 &7.8 &n/a &19.6&19.7&20.3 \\
         \hline

    \end{tabular}
    }
    \caption{Sensitivity of different specificity metrics to the strength of the specificity regularizer. The model is Qwen2.5-7B-Instruct, and the dataset is MCF.
}
    \label{tab: correlation with specificity mcf qwen}
\end{table*}

\begin{table*}[]
    \centering
    \resizebox{1.99\columnwidth}{!}{
    \begin{tabular}{c|c |c |c |c |c |c |c |c |c|c|c |c  }
    \hline
      \multirow{2}{*}{\diagbox{Methods}{Metrics}}   & \multicolumn{6}{c|}{  Specificity regularizer  $||\Delta k_j|| $ (mean)}& \multicolumn{2}{c|}{ GT-based specificity} & \multicolumn{4}{c}{ GT-free specificity  }\\
      \cline{2-13}
         & Layer 4 & Layer 5 & Layer 6 & Layer 7 & Layer 8&Average & S-acc $\uparrow$ & T-acc $\uparrow$& $D_{KL}$ $\downarrow$& Top-1 $\uparrow$&  Top-5 $\uparrow$& Top-10 $\uparrow$\\
         \hline
          pre-edit&0&0&0&0&0&0&0.0&38.6&0&100&100&100\\
         $\lambda=1.5\times10^6$&0.050&0.080&0.146&0.227&0.746&0.250&0.3&41.1&0.918&49.5&62.9&67.2\\
         $\lambda=1.5\times10^5$&1.326&1.904&3.436&5.837&17.438&5.99&5.8&46.5&3.943&17.0&27.7&32.5\\
         $\lambda=1.5\times10^4$&11.130&15.723&22.793&42.034&95.020&37.340&6.4&45.8&4.783&13.2&23.4&27.5\\
         $\lambda=1.5\times10^3$&64.446&83.313&83.003&101.953&91.893&84.922&4&39.1&5.018&5.6&17.0&21.8\\
         $\lambda=1.5\times10^2$&396.573&112.731&53.134&70.777&57.640&138.171&0.8&23.9&4.755&4.5&16.0&21.7\\
         \hline
         \\
         \hline
         Kendall’s $\tau$ correlation $\uparrow$ &1.00 &1.00 &0.87&0.87 &0.60& n/a &0.33 &0.20 &0.73 & 1.00&1.00&1.00 \\
         Range & n/a& n/a& n/a& n/a& n/a& n/a&0 $\sim$ 100 &0 $\sim$ 100 &n/a &0 $\sim$ 100&0 $\sim$ 100&0 $\sim$ 100
         \\
         (Max-Min)/Range $\uparrow$&  n/a& n/a& n/a& n/a& n/a& n/a&0.064 &0.226 &n/a &0.955&0.840&0.783 \\
         Standard deviation $\uparrow$ &  n/a& n/a& n/a& n/a& n/a& n/a&2.9 &8.2 &n/a &37.3&33.6&31.8 \\
         \hline
    \end{tabular}
    }
    \caption{Sensitivity of different specificity metrics to the strength of the specificity regularizer. The model is Qwen2.5-7B-Instruct, and the dataset is ZsRE. 
}
    \label{tab: correlation with specificity zsre qwen}
\end{table*}

\begin{table*}[t]
    \centering
    
    \resizebox{1.99\columnwidth}{!}{

    \begin{tabular}{c| c |c |c |c |c |c |c |c  }
    \hline
     \multicolumn{9}{c}{ The MCF dataset}
    \\
    \hline
    
      \multicolumn{1}{c|}{\multirow{2}{*}{\diagbox{Methods}{Metrics}}}   & \multicolumn{2}{c|}{ Target knowledge}& \multicolumn{2}{c|}{ GT-based specificity} & \multicolumn{4}{c}{ GT-free specificity  }\\
      \cline{2-9}
         \multicolumn{1}{c|}{} & Efficacy $\uparrow$& Generalization $\uparrow$& S-acc $\uparrow$ & C-acc $\uparrow$& $D_{KL}$ $\downarrow$& Top-1 $\uparrow$&  Top-5 $\uparrow$& Top-10 $\uparrow$\\
         \hline
          \multicolumn{1}{c|}{pre-edit}&21.7& 24.1&10.5 &77.8&0 &100 & 100 &100\\
          \multicolumn{1}{c|}{WISE} & 24.8 & 26.6 & 9.9 &75.6 & 0.22 & 80.1 & 74.9 & 74.0\\
          \multicolumn{1}{c|}{MEMIT}& 94.0 & 82.1& 12.0& 73.3 & 0.81& 63.5&60.8&59.4\\
          \multicolumn{1}{c|}{Adaedit }& 94.7 & 87.5 &9.7&67.2& 1.92 & 46.5 & 46.2 &44.6\\
          \multicolumn{1}{c|}{Alphaedit} & 99.4& 99.1& 7.4&68.6& 0.98&62.9&57.8&55.9\\
          \multicolumn{1}{c|}{EMMET} & 99.3& 91.8&9.4& 69.8&1.02&59.1&55.9&54.2\\
          \multicolumn{1}{c|}{NAMET} & 76.8&62.2&9.6&72.7&0.52& 71.4&64.4&62.3\\
          \multicolumn{1}{c|}{PMET} & 95.3& 88.5 & 9.8&66.8 & 2.02 & 45.3 &45.1 &43.6\\ 
          \multicolumn{1}{c|}{PRUNE} & 93.8 & 82.2 & 12.1 &73.2 &0.80 & 63.5 & 60.9 &59.5\\
          \multicolumn{1}{c|}{RECT }& 93.4 & 81.5 &12.1 &73.4 & 0.80 &63.7 & 61.8 &59.7\\
          \hline
\multicolumn{9}{c}{}
\\
\hline
     \multicolumn{9}{c}{ The ZsRE dataset}
    \\
    \hline
               \multicolumn{1}{c|}{\multirow{2}{*}{\diagbox{Methods}}{Metrics}}   & \multicolumn{2}{c|}{ Target knowledge}& \multicolumn{2}{c|}{ GT-based specificity} & \multicolumn{4}{c}{ GT-free specificity  }\\
      \cline{2-9}
         \multicolumn{1}{c|}{}& Efficacy $\uparrow$& Generalization $\uparrow$& S-acc $\uparrow$ & T-acc $\uparrow$& $D_{KL}$ $\downarrow$& Top-1 $\uparrow$&  Top-5 $\uparrow$& Top-10 $\uparrow$\\
         \hline
          pre-edit& 23.7 & 22.7& 0 & 24.9 &0 &100 &100 &100\\
          WISE & 9.1 & 9.2 & 0 & 8.2 & 2.48 & 0.8 & 4.8 & 10.5\\
          MEMIT& 81.4 & 74.4 &0.4 &26.5 & 0.90 &70.6 & 47.1 &49.5\\
          Adaedit & 82.3 & 78.2 & 0.7 &24.9 & 2.89 &24.6 &19.6 &21.0\\
          Alphaedit & 92.0 & 83.2 & 0.1 &25.2 & 0.93 & 73.0 &45.0 & 48.9\\
          EMMET & 90.9 &82.5 &0.1 &25.7 &1.06 &67.1 & 43.4 & 45.8\\
          NAMET & 40.6 & 37.5 & 0 &24.5 &0.52 &84.6 &55.3 &57.5\\
          PMET & 82.9 & 79.0 &0.6 &25.0 & 3.02 &22.3 &18.4 &20.2\\ 
          PRUNE & 81.4 & 74.3 & 0.4& 26.4 & 0.90 &70.5 &47.2 &49.6\\
          RECT& 80.7 & 73.5 &0.4 &26.5 & 0.89 &70.7 &47.5 &49.8\\
          
             \hline
\multicolumn{9}{c}{}
\\
\hline
     \multicolumn{9}{c}{ The rank stability (Kendall's $\tau$) of different methods across datasets}
    \\
    \hline
               \multirow{2}{*}{Metrics}   & \multicolumn{2}{c|}{ Target knowledge}& \multicolumn{2}{c|}{ GT-based specificity} & \multicolumn{4}{c}{ GT-free specificity  }\\
      \cline{2-9}
         \multicolumn{1}{c|}{}& Efficacy $\uparrow$& Generalization $\uparrow$& S-acc $\uparrow$ & T-acc $\uparrow$& $D_{KL}$ $\uparrow$& Top-1 $\uparrow$&  Top-5 $\uparrow$& Top-10 $\uparrow$\\
    \hline    
    & 0.94 &0.91 &0.1 &0.11 &0.73 &0.47 &0.64 &0.64 \\
    \hline
    Average &  \multicolumn{2}{c|}{0.93}& \multicolumn{2}{c|}{0.11} & \multicolumn{4}{c}{ 0.62 }\\
    \hline

    \end{tabular}
    }
    \caption{Benchmarking different editing methods. The model is GPT2-XL.
}
    \label{tab: benchmark gpt2-xl}
\end{table*}

\begin{table*}[t]
    \centering
    \resizebox{1.99\columnwidth}{!}{
    \begin{tabular}{c| c |c |c |c |c |c |c |c  }
    \hline
     \multicolumn{9}{c}{ The MCF dataset}
    \\
    \hline
    
      \multicolumn{1}{c|}{\multirow{2}{*}{\diagbox{Methods}{Metrics}}}   & \multicolumn{2}{c|}{ Target knowledge}& \multicolumn{2}{c|}{ GT-based specificity} & \multicolumn{4}{c}{ GT-free specificity  }\\
      \cline{2-9}
         \multicolumn{1}{c|}{} & Efficacy $\uparrow$& Generalization $\uparrow$& S-acc $\uparrow$ & C-acc $\uparrow$& $D_{KL}$ $\downarrow$& Top-1 $\uparrow$&  Top-5 $\uparrow$& Top-10 $\uparrow$\\
         \hline
          \multicolumn{1}{c|}{pre-edit}&13.7& 16.5&14.7 &85.9&0 &100 & 100 &100\\
          \multicolumn{1}{c|}{WISE} & 13.8 & 16.7 & 14.6 &85.7 & 0.04 & 99.4 & 99.4 & 99.5\\
          \multicolumn{1}{c|}{MEMIT}& 99.9 & 98.4& 9.8& 73.9 & 1.03& 60.6&60.8&59.6\\
          \multicolumn{1}{c|}{Adaedit }& 99.0 & 97.6 &8.2&68.5& 1.55 & 52.8 & 51.5 &49.8\\
          \multicolumn{1}{c|}{Alphaedit} & 99.4& 98.0& 6.2&67.3& 1.58&52.0&52.4&50.5\\
          \multicolumn{1}{c|}{EMMET} & 99.8& 97.9&7.6& 68.7&1.39&54.2&54.0&52.6\\
          \multicolumn{1}{c|}{NAMET} & 99.9&98.6&9.7&74.0&1.01& 60.9&60.8&59.9\\
          \multicolumn{1}{c|}{PMET} & 98.9& 97.2 & 8.4&68.0 & 1.56 & 52.0 &51.0 &49.6\\ 
          \multicolumn{1}{c|}{PRUNE} & 99.8 & 98.5 & 10.1 &74.3 &0.99 & 61.3 & 61.3 &60.3\\
          \multicolumn{1}{c|}{RECT }& 99.8 & 98.4 &10.0 &74.2 & 1.01 &60.5 & 60.9 &60.0\\
          \hline
\multicolumn{9}{c}{}
\\
\hline
     \multicolumn{9}{c}{ The ZsRE dataset}
    \\
    \hline
               \multicolumn{1}{c|}{\multirow{2}{*}{\diagbox{Methods}}{Metrics}}   & \multicolumn{2}{c|}{ Target knowledge}& \multicolumn{2}{c|}{ GT-based specificity} & \multicolumn{4}{c}{ GT-free specificity  }\\
      \cline{2-9}
         \multicolumn{1}{c|}{}& Efficacy $\uparrow$& Generalization $\uparrow$& S-acc $\uparrow$ & T-acc $\uparrow$& $D_{KL}$ $\downarrow$& Top-1 $\uparrow$&  Top-5 $\uparrow$& Top-10 $\uparrow$\\
         \hline
          pre-edit& 36.4 & 35.4& 0.4 & 38.4 &0 &100 &100 &100\\
          WISE & 35.7 & 34.8 & 0 & 33.1 & 0.39 & 94.6 & 94.6 & 94.6\\
          MEMIT& 93.7 & 90.5 &4.6 &45.6 & 4.16 &15.0 & 26.6 &31.1\\
          Adaedit & 86.2 & 83.8 & 6.5 &45.5 & 5.15 &5.9 &16.4 &20.4\\
          Alphaedit & 82.8 & 79.9 & 2.9 &37.5 & 4.50 & 11.5 &22.1 & 25.9\\
          EMMET & 84.7 &80.6 &5.0 &39.9 &4.86 &5.7 & 17.8 & 21.9\\
          NAMET & 93.9 & 90.7 & 4.2 &45.1 &4.00 &15.9 &27.4 &32.0\\
          PMET & 86.1 & 83.2 &6.0 &45.1 & 5.06 &5.9 &16.5 &20.8\\ 
          PRUNE & 94.0 & 90.7 & 4.4& 45.6 & 4.21 &15.2 &26.9 &31.1\\
          RECT& 93.7 & 90.6 &4.4 &45.6 & 4.20 &14.8 &26.7 &30.9\\
          
             \hline
\multicolumn{9}{c}{}
\\
\hline
     \multicolumn{9}{c}{ The rank stability (Kendall's $\tau$) of different methods across datasets}
    \\
    \hline
               \multirow{2}{*}{Metrics}   & \multicolumn{2}{c|}{ Target knowledge}& \multicolumn{2}{c|}{ GT-based specificity} & \multicolumn{4}{c}{ GT-free specificity  }\\
      \cline{2-9}
         \multicolumn{1}{c|}{}& Efficacy $\uparrow$& Generalization $\uparrow$& S-acc $\uparrow$ & T-acc $\uparrow$& $D_{KL}$ $\uparrow$& Top-1 $\uparrow$&  Top-5 $\uparrow$& Top-10 $\uparrow$\\
    \hline    
    & 0.61 &0.70&-0.43&0.07&0.6&0.73&0.78&0.78\\
    \hline
    Average &  \multicolumn{2}{c|}{0.66}& \multicolumn{2}{c|}{-0.18} & \multicolumn{4}{c}{ 0.72 }\\
    \hline

    \end{tabular}
    }
    \caption{Benchmarking different editing methods. The model is Qwen2.5-7B-Instruct.
}
    \label{tab: benchmark mcf qwen}
\end{table*}

\end{document}